\documentclass[a4paper,twoside]{article}

\usepackage{epsfig}
\usepackage{subcaption}
\usepackage{calc}
\usepackage{amssymb}
\usepackage{amstext}
\usepackage{amsmath}
\usepackage{amsthm}
\usepackage{multicol}
\usepackage{pslatex}
\usepackage{apalike}
\usepackage{algorithm2e}
\usepackage[bottom]{footmisc}
\usepackage{SCITEPRESS}     

\usepackage{multirow}
\usepackage{amsmath}
\usepackage{etoolbox} 
\usepackage{placeins} 
\usepackage{amsmath}
\usepackage{booktabs} 

\begin{document}

\title{Classifier Ensemble for Efficient Uncertainty Calibration of Deep Neural Networks for Image Classification}

\author{\authorname{
Michael Schulze\sup{1},
Nikolas Ebert\sup{1},
Laurenz Reichardt\sup{1},
and Oliver Wasenm\"uller\sup{1}}
\affiliation{\sup{1}Mannheim University of Applied Sciences, Germany}
\email{\{m.schulze, n.ebert, l.reichardt, o.wasenmueller\}@hs-mannheim.de
}
}


\abstract{This paper investigates novel classifier ensemble techniques for uncertainty calibration applied to various deep neural networks for image classification. We evaluate both accuracy and calibration metrics, focusing on Expected Calibration Error (ECE) and Maximum Calibration Error (MCE). Our work compares different methods for building simple yet efficient classifier ensembles, including majority voting and several metamodel-based approaches. Our evaluation reveals that while state-of-the-art deep neural networks for image classification achieve high accuracy on standard datasets, they frequently suffer from significant calibration errors. Basic ensemble techniques like majority voting provide modest improvements, while metamodel-based ensembles consistently reduce ECE and MCE across all architectures. Notably, the largest of our compared metamodels demonstrate the most substantial calibration improvements, with minimal impact on accuracy. Moreover, classifier ensembles with metamodels outperform traditional model ensembles in calibration performance, while requiring significantly fewer parameters. In comparison to traditional post-hoc calibration methods, our approach removes the need for a separate calibration dataset. These findings underscore the potential of our proposed metamodel-based classifier ensembles as an efficient and effective approach to improving model calibration, thereby contributing to more reliable deep learning systems.}

\onecolumn \maketitle \normalsize \setcounter{footnote}{0} \vfill

\section{\uppercase{Introduction}}
\label{sec:introduction}

Machine learning models, particularly deep neural networks, are increasingly applied in safety critical areas such as autonomous driving \cite{ebert2022multitask,reichardt2023360} and medical image analysis \cite{ebert2023transformer}, where incorrect decisions can have serious consequences. In these settings, achieving high accuracy and robustness \cite{oehri2024genformer,kendall2017uncertainties} is crucial, but models must also provide reliable uncertainty estimates to assess whether their predictions can be trusted \cite{jiang2018trust}. Calibration addresses this need by aligning predicted probabilities with the true likelihood of predictions being correct \cite{brocker2009reliability}. However many machine learning models \cite{niculescu2005predicting}, especially deep neural networks \cite{guo2017calibration}, are poorly calibrated and tend to produce overconfident predictions, even when they are wrong.

\begin{figure}[t]
  \centering
  \includegraphics[width=0.93\linewidth]{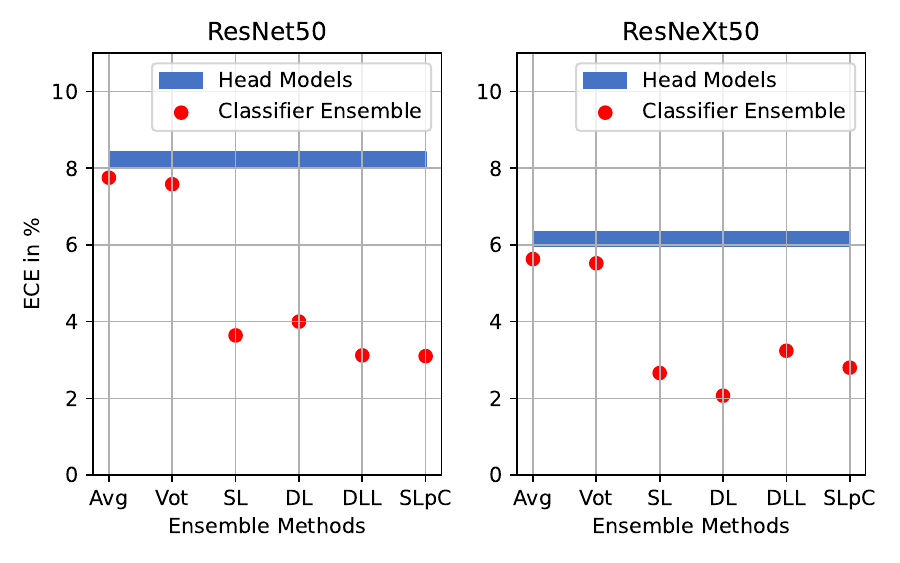}
  \caption{Expected Calibration Error (ECE) of ResNet50 \cite{he2016deep} (left) and ResNeXt50 \cite{xie2017aggregated} (right) on CIFAR-100 \cite{krizhevsky2009learning}. Each model was trained with five classifier heads initialized with different random seeds but using the same backbone. The blue area represents the ECE range for the uncalibrated classifiers. Each red dot corresponds to the ECE value achieved using different ensemble techniques. The use of metamodels (SL, DL, DLL, SLpC) significantly improves the calibration performance and reduces the ECE compared to the uncalibrated baseline.
}
  \label{fig:teaser}
\end{figure}

Post-hoc calibration methods, which adjust the prediction scores of a trained neural network using a separate calibration dataset, are widely used to improve uncertainty estimates. Examples include Platt scaling \cite{platt1999probabilistic}, histogram binning \cite{zadrozny2001learning}, isotonic regression \cite{zadrozny2002transforming} and temperature scaling \cite{guo2017calibration}. Parametric methods like temperature scaling rescale the output logits of a neural network for classification using learned parameters from a calibration set. However, in many real-world scenarios with limited data, a dedicated calibration set is not available.
Although non-parametric methods, such as isotonic regression, offer greater flexibility, they can reduce model accuracy after calibration. Similar to their parametric counterparts, these methods also require a dedicated calibration set.

\begin{figure}[t]
  \centering
  \includegraphics[width=0.90\linewidth]{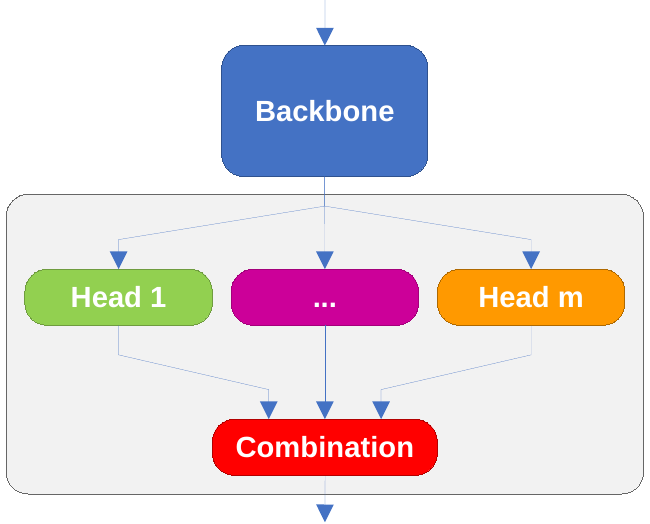}
  \caption{The principle of the classifier ensemble involves a single backbone that feeds multiple classifiers (heads). The combination method can be freely selected.}
  \label{fig:classifier-ensemble}
\end{figure}

In contrast to post-hoc calibration, ab-initio methods \cite{lakshminarayanan2017simple,kumar2018trainable} aim to train models that are well-calibrated from the start, incorporating uncertainty directly during training. 
Furthermore, deep ensembles \cite{lakshminarayanan2017simple,wenzel2020hyperparameter} combine multiple models trained on the same dataset through majority voting or averaging, which enhances accuracy and reduces uncertainty. However, a disadvantage of this approach is the high computational cost associated with training several independent models. In contrast, Monte Carlo dropout \cite{gal2016dropout} follows a similar strategy by applying dropout during training and inference to randomly deactivate individual neurons, thereby creating an ensemble of models. However, this method requires repeated inference, resulting in lower accuracy and higher uncertainty compared to deep ensembles.

Thus, we propose a novel approach based on classifier ensemble (see Figure \ref{fig:classifier-ensemble}), which effectively combines transfer learning with ensemble methods for efficient uncertainty calibration. 
In contrast to traditional ensemble techniques, where multiple full-scale networks are trained separately and their predictions are combined, our method focuses on training multiple lightweight classifiers on-top of a shared backbone and utilizing their predictions collaboratively.
This technique stands out by eliminating the need for an additional calibration dataset and significantly reducing computational overhead during both training and inference. By combining the strengths of transfer learning and ensemble methods, our classifier ensemble significantly reduces uncertainty while maintaining computational efficiency. Furthermore, we have proven the effectiveness of our approach in numerous analyses of different neural networks (see Figure \ref{fig:teaser}) on CIFAR-100 \cite{krizhevsky2009learning} and TinyImageNet \cite{le2015tiny} benchmarks from the field of image classification.

\section{\uppercase{Related Work}}
\label{sec:rw}

\subsection{Calibration Methods}
During the past decade, several post-hoc methods for calibrating network outputs have been developed. Histogram binning \cite{zadrozny2001learning} assigns predictions to fixed intervals and learns a calibrated score for each by minimizing the squared error loss on a calibration dataset. During inference, uncalibrated scores are replaced by these calibrated scores.
Isotonic regression \cite{zadrozny2002transforming} generalizes this method by dynamically learning intervals from the calibration dataset, adjusting both boundaries and calibrated scores to produce a piecewise constant function. Logistic regression, or Platt scaling \cite{platt1999probabilistic}, uses uncalibrated scores as features for a regression model trained to minimize negative log-likelihood, which then calibrates the scores during prediction.
Similar to Platt scaling, temperature scaling (TS) \cite{guo2017calibration} uses a single scalar parameter to adjust the prediction scores based on a calibration dataset, preserving model accuracy. An extension of TS called Ensemble Temperature Scaling \cite{zhang2020mix} learns a mapping of three scaling factors instead of a single factor, resulting in a weighted combination of three TS. Parameterized Temperature Scaling \cite{tomani2022parameterized} extends TS by using a small neural network to learn multiple parameters for different classes instead of a single parameter for all classes.

In contrast to the mentioned post-hoc methods, deep ensembles \cite{lakshminarayanan2017simple} involve the training of multiple models on the same dataset and combining them through majority voting or averaging, enhancing accuracy and reducing uncertainty. However, this requires significant computational resources.
Monte Carlo dropout \cite{gal2016dropout} combines predictions from different subnetworks by applying dropout during training and inference. This method generates an ensemble by performing multiple inferences with different active neurons, but it generally results in lower accuracy and higher uncertainty compared to deep ensembles.

\subsection{Model Ensemble}
Model ensemble techniques combine multiple individual models to enhance predictive performance. The core idea is that different models may possess unique strengths and weaknesses, which can be maximized and balanced through aggregation, leading to improved overall accuracy.
In a voting ensemble \cite{goodfellow2016deep}, several models are trained on the same dataset, and their predictions are aggregated through majority voting. This approach effectively utilizes the collective intelligence of the models and is suitable when individual models exhibit similar performance levels but make distinct errors.
Deep ensemble \cite{lakshminarayanan2017simple,wenzel2020hyperparameter} methods involve independent training multiple neural networks, each with its own weights and parameters. Their predictions are aggregated via averaging or majority voting, capturing diverse aspects of the data and yielding more robust predictions.
Bagging ensembles \cite{raschka2022machine} use bootstrapping to create multiple subsets from the training data by drawing random samples with replacement. Models are trained on these subsets, and their predictions are combined through averaging or voting. This method reduces model variance and enhances robustness against overfitting.
In boosting ensembles \cite{raschka2022machine}, several weak models are trained sequentially, and their predictions are combined through weighted averaging. The weights are adjusted to emphasize samples that previous models misclassified, addressing issues of high bias or underfitting.
Stacking ensembles \cite{raschka2022machine} involve training multiple models on the same dataset and using their predictions as features for a meta-model. The meta-model is trained on the predictions of the base models with true labels as targets, allowing for the integration of diverse strengths and weaknesses to enhance predictive accuracy.

Unlike the ensemble methods mentioned above, we do not rely on retaining multiple full-scale networks. Instead, we retrain multiple lightweight classifiers (each comprising less than 1\% of the entire model) with a strong shared backbone and utilize their predictions collaboratively. This approach effectively reduces model uncertainty and yields a well-calibrated model without the need for a dedicated calibration dataset, which is typically required by other post-hoc methods.

\section{\uppercase{Method}}
\label{sec:Method}

\subsection{Preliminaries}
Let \( X \in \mathbb{R}^D \) represent the D-dimensional input and \( Y \in \{1, \dots, \mathcal{C}\} \) represent the class labels for a classification task with \( \mathcal{C}  \) possible classes. The joint distribution of \( X \) and \( Y \) is denoted by \( \pi(X, Y) = \pi(Y|X)\pi(X) \). The dataset \( \mathcal{D} \) consists of \( N \) independent and identically distributed (i.i.d.) samples \( \mathcal{D} = \{(X_n, Y_n)\}_{n=1}^N \), drawn from this distribution. A neural network classifier \( h(X) \) outputs a predicted class \( \hat{Y} \) and a corresponding logit vector \( \hat{Z} \). The logits \( \hat{Z} \) are then converted into a confidence score \( \hat{P} \) for the predicted class \( \hat{Y} \) using the softmax function \( \sigma_{\text{SM}} \), where \( \hat{P} = \max_c \sigma_{\text{SM}}(\hat{Z})_c \).

\textbf{Uncertainty calibration:}
Perfect calibration is defined as the condition where the accuracy of predictions aligns with the confidence levels across all possible confidence values \cite{guo2017calibration}, mathematically represented as 
\begin{equation}
    \mathbb{P}(\hat{Y} = Y | \hat{P} = p) = p \quad \text{for every } p \in [0, 1].
\end{equation}
In contrast, miss-calibration refers to the expected discrepancy between confidence and accuracy, which can be expressed as:
\begin{equation}\label{eq:error}
    \mathbb{E}_{\hat{P}} \left[|\mathbb{P}(\hat{Y} = Y | \hat{P} = p) - p |\right].
\end{equation}

\textbf{Measuring uncertainty:}
The Expected Calibration Error (ECE) serves as a widely used scalar metric for assessing miss-calibration \cite{naeini2015obtaining}. It approximates Equation (\ref{eq:error}) based on the predictions \( \hat{Y} \), the confidence scores \( \hat{P} \) and the ground truth labels \( Y \) of a finite number of \( N \) samples. The ECE is computed by dividing the confidence scores into \( M \) equal bins \(B_m\), calculating the average confidence (conf) and classification accuracy (acc) for each bin, and then summarizing the resulting differences. The formula for ECE is:
\begin{equation}
    ECE^d = \sum_{m=1}^{M} \frac{|B_m|}{N} \|\text{acc}(B_m) - \text{conf}(B_m)\|_d,
\end{equation}
where \( d \) is typically set to 1 for the L1-norm.  

In addition to ECE, we use the Maximum Calibration Error (MCE), which captures the largest discrepancy among the intervals used to calculate the ECE, providing another measure of calibration performance. The formula for MCE is:

\begin{equation}
    \label{eq:mce}
    {MCE} = \max\limits_{m\in\lbrace 1,\ldots,M \rbrace} \lvert \text{{acc}}(B_m) - \text{{conf}}(B_m) \rvert
\end{equation}

\subsection{Classifier Ensemble for Uncertainty Calibration}
A commonly used method for calibrating model outputs is deep ensembles \cite{lakshminarayanan2017simple} (see Section \ref{sec:rw}), where multiple models are trained on the same data and combined during inference. However, this approach requires substantial time and computational resources, as it necessitates training several models from scratch and performing multiple inferences.

In contrast, our novel classifier ensemble approach divides the model into a backbone and a head (classifier), with the backbone responsible for computing features and being significantly larger than the head, which maps these features to target classes. Notably, we only re-train the heads while keeping the pre-trained backbone frozen. The individual classifiers are subsequently combined using model ensemble techniques, as illustrated in Figure \ref{fig:classifier-ensemble}.

\subsubsection{Train Strategies}

\begin{figure}[t]
\centering
   \includegraphics[width=\linewidth]{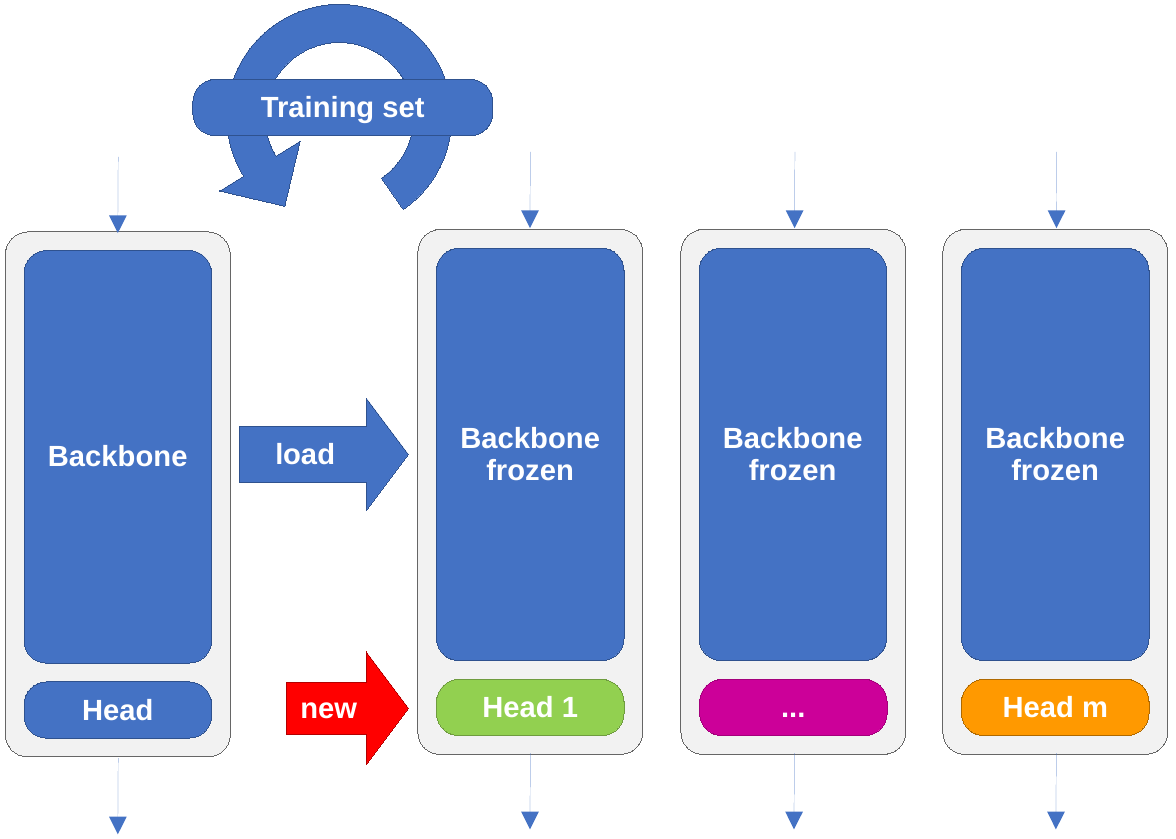}
    \caption{Training Process of our classifier ensemble. }
    \label{fig:ensemble-training}
\end{figure}

The training of a classifier ensemble is conducted in multiple steps. Initially, a base model is created and trained on the training dataset, after which it is saved. Subsequently, a new base model is created, and the weights from the previously trained model are loaded. Following the principles of transfer learning, the weights are frozen, and only the head is newly constructed and then trained again on the training data. This process is repeated as many times as necessary to form the desired number of heads for the classifier ensemble, as illustrated in Figure \ref{fig:ensemble-training}.

Such a separate training approach offers several advantages. It allows the use of different head architectures, such as varying the number of layers or incorporating dropout. Additionally, diverse data augmentation strategies or different subsets of the dataset can be applied during each head's training, akin to the bagging ensemble method described in Section \ref{sec:rw}.

\subsubsection{Ensemble Methods}

In the final step of the classifier ensemble, the different heads must be combined, as shown in Figure \ref{fig:classifier-ensemble}. Our proposed methods for combining these heads include averaging, voting, and the use of metamodels. Averaging involves summing the individual outputs of the heads and dividing the total by the number of heads. In voting, a majority decision is made by selecting the most frequent predicted values across the heads. 

Alternatively, metamodels can be used, where the classifiers are combined using additional learnable parameters. One approach involves concatenating the outputs of all \( m \) heads and applying a fully connected layer, where the input consists of the combined predictions from the heads, yielding \( m \cdot \mathcal{C} \) input features, while the output remains the original \( \mathcal{C} \) classes.

The architecture can be further extended with additional hidden layers, nonlinearities, or dropout, as long as the structure supports \( m \cdot \mathcal{C} \) input and \( \mathcal{C} \) output features. Another variant is to link the head outputs class-wise with a fully connected layer. In this case, a separate fully connected layer is used for each class, with each layer having \( m \) input features and a single output, leading to a total of \( \mathcal{C} \) fully connected layers, one for each class.

In the studies conducted in Section \ref{sec:results}, we performed a thorough comparison of all methods. However, no single approach consistently outperformed the others across different networks and datasets. Nevertheless, all methods demonstrated a significant improvement compared to the uncalibrated baseline.

\section{\uppercase{Evaluation}}
\label{sec:eval}
Our evaluation is divided into three sections. Section \ref{sec:data} first provides a detailed overview of the data and training settings used for all our experiments. Next, in section \ref{sec:cifar}, we conduct an extensive study with CIFAR-100 \cite{krizhevsky2009learning}. Finally, in Section \ref{sec:tiny}, we use Tiny ImageNet \cite{le2015tiny} for further evaluations.

\begin{table*}[ht]
  \caption{Comparison of accuracy, ECE and MCE in percent of uncalibrated heads and their combination to the classifier ensemble with ResNet \cite{he2016deep} variants on CIFAR-100 \cite{krizhevsky2009learning}.}
  \label{tab:vergleich-classifier-ensemble-teil1-cifar100}
  \renewcommand{\arraystretch}{1.2} 
  \centering
  \sffamily
  \begin{footnotesize}
    \begin{tabular}{llrrr|rrr|rrr|rrr}
    \toprule
\multirow{2}{*}{} & \multirow{2}{*}{Model} & \multicolumn{3}{c|}{ResNet18} & \multicolumn{3}{c|}{ResNet34} & \multicolumn{3}{c|}{ResNet50} & \multicolumn{3}{c}{ResNet101} \\
 & & Acc. & ECE & MCE & Acc. & ECE & MCE & Acc. & ECE & MCE & Acc. & ECE & MCE \\
    \midrule
\multirow{5}{*}{\rotatebox[origin=c]{90}{Baseline}} & Head 1 & 75.08 & 4.41 & 27.47 & 76.74 & 5.60 & 15.66 & 77.21 & 8.22 & 25.01 & 78.06 & 8.76 & 21.06 \\
 & Head 2 & 74.95 & 4.69 & 16.26 & 76.76 & 5.74 & 15.00 & 77.15 & 8.30 & 23.80 & 78.06 & 8.83 & 23.14 \\
 & Head 3 & 75.07 & 4.65 & 27.25 & 76.86 & 5.56 & 27.76 & 77.45 & 8.06 & 23.73 & 78.12 & 8.68 & 22.40 \\
 & Head 4 & 75.22 & 4.40 & 24.07 & 76.89 & 5.92 & 17.64 & 77.39 & 8.04 & 25.96 & 78.20 & 8.64 & 23.26 \\
 & Head 5 & 75.25 & 4.46 & 14.11 & 76.71 & 5.75 & 15.20 & 77.06 & 8.43 & 25.51 & 78.05 & 8.76 & 23.11 \\
    \hline
\multirow{6}{*}{\rotatebox[origin=c]{90}{\begin{tabular}{c}Classifier \\ Ensemble (ours)\end{tabular}}} & Avg. & 75.06 & 4.43 & 10.66 & 76.83 & 5.75 & 19.12 & 77.28 & 7.75 & 25.13 & 78.07 & 8.43 & 20.17 \\
 & Vot. & 74.96 & 4.25 & 11.97 & 76.81 & 5.12 & 14.27 & 77.29 & 7.58 & 24.97 & 77.99 & 8.31 & 22.85 \\
  \cline{2-14} 
 & SL & 74.39 & \textbf{2.59} & \textbf{8.35} & 76.45 & 3.48 & 11.37 & 77.17 & 3.64 & 9.62 & 77.30 & 2.81 & 8.61 \\
 & DL & 74.29 & 2.93 & 8.44 & 76.37 & \textbf{3.44} & 9.10 & 76.89 & 4.00 & 10.52 & 77.44 & \textbf{2.71} & \textbf{7.57} \\
 & DLL & 74.73 & 3.51 & 10.22 & 76.75 & 3.83 & 11.12 & 77.32 & 3.12 & 11.10 & 77.94 & 3.39 & 10.66 \\
 & SLpC & 74.99 & 4.11 & 11.71 & 76.73 & 4.01 & \textbf{7.73} & 77.11 & \textbf{3.10} & \textbf{9.35} & 78.22 & 3.70 & 9.12 \\
    \bottomrule
    \end{tabular}
  \end{footnotesize}
  \rmfamily
\end{table*}

\subsection{Datasets, Training, and Ensemble Configuration}\label{sec:data}
\textbf{Datasets:} To evaluate our novel classifier ensemble, we utilize the CIFAR-100 \cite{krizhevsky2009learning} dataset. CIFAR-100 consists of 50,000 training images and 10,000 test images with 100 classes. In addition to the experiments on the CIFAR-100 dataset, we also conducted experiments on the Tiny ImageNet \cite{le2015tiny} dataset consisting of 100,000 training images and 5,000 test images of 200 classes. 

\textbf{Base Training:} As outlined in Section \ref{sec:Method}, the first step in training our classifier ensemble is the standard pre-training of a base model (backbone + head), which serves as the foundation for subsequent steps. For this work, a Stochastic Gradient Descent (SGD) optimizer with momentum of 0.9 and weight decay of 5e-04 is used. During the 200 training epochs, the basic learning rate of 0.1 is gradually adjusted by a factor of 0.2 using a multi-stage scheduler. A batch size of 128 and a basic data augmentation strategy, including random cropping, padding, horizontal flipping and random rotation is used.

\textbf{Training Heads:} The individual heads are created by loading the base model. As described in Section \ref{sec:Method}, the backbone weights are frozen and only the classifier (head) is reinitialized with new random seeds. The new classifier is then trained using an SGD optimizer with an initial learning rate of 0.1. The learning rate for training the heads is adjusted using a Plateau-Min-Scheduler, which monitors the validation loss. If the loss does not improve within a specified number of epochs, the learning rate is reduced by multiplying it by a factor of 0.5. Additionally, early stopping with a patience of 15 epochs is applied, terminating the training if no further improvements are observed.

All heads used in this work consist of a single fully connected layer. Each base model is trained with five distinct heads, which are saved and later combined into an ensemble. Since each head contains only a few parameters, the training process is very fast.

\textbf{Ensemble Configuration:} For classifier ensemble without a metamodel, two combination methods were explored: mean averaging and majority voting. When using a metamodel to combine the heads, additional training is required. Four metamodels were implemented and analyzed: Single-Layer (SL), Double-Layer (DL), Double-Layer-Large (DLL), and Single-Layer-per-Class (SLpC).

The SL metamodel combines the outputs of the heads through a single fully connected layer. The DL metamodel adds a second layer with ReLU activation and dropout, where the first layer reduces the number of neurons. In contrast, the DLL metamodel doubles the number of neurons in the first layer compared to the DL model. The SLpC metamodel takes a different approach, using a dedicated fully connected layer for each class, where the concatenated head outputs are connected with class-specific layers.

As the metamodels introduce additional parameters, they require training on the training dataset. This training is performed over 20 epochs using an SGD optimizer with an initial learning rate of 0.0002, along with a Plateau-Min-Scheduler to adjust the learning rate. After training, the metamodel with the lowest validation loss is selected for deployment.

\subsection{Results}\label{sec:results}

\begin{table*}[ht]
  \caption{Comparison of accuracy, ECE and MCE in percent of uncalibrated heads and their combination to the classifier ensemble with ResNeXt \cite{xie2017aggregated}, DenseNet \cite{huang2017densely} and GoogLeNet \cite{szegedy2015going} on CIFAR-100 \cite{krizhevsky2009learning}.}
  \label{tab:vergleich-classifier-ensemble-teil2-cifar100}
  \renewcommand{\arraystretch}{1.2} 
  \centering
  \sffamily
  \begin{footnotesize}
    \begin{tabular}{llrrr|rrr|rrr|rrr}
    \toprule
\multirow{2}{*}{} & \multirow{2}{*}{Model} & \multicolumn{3}{c|}{ResNeXt50} & \multicolumn{3}{c|}{DenseNet121} & \multicolumn{3}{c|}{DenseNet169} & \multicolumn{3}{c}{GoogLeNet} \\
 & & Acc. & ECE & MCE & Acc. & ECE & MCE & Acc. & ECE & MCE & Acc. & ECE & MCE \\
    \midrule
\multirow{5}{*}{\rotatebox[origin=c]{90}{Baseline}} & Head 1 & 77.15 & 6.05 & 15.06 & 77.55 & 4.74 & 10.20 & 78.43 & 4.00 & 10.28 & 75.66 & 6.93 & 16.65 \\
 & Head 2 & 77.06 & 5.99 & 12.28 & 77.45 & 4.75 & 12.02 & 78.56 & 3.92 & 9.30 & 75.84 & 7.02 & 19.36 \\
 & Head 3 & 76.86 & 6.14 & 13.56 & 77.55 & 4.97 & 13.04 & 78.57 & 4.08 & 9.64 & 75.74 & 6.96 & 18.68 \\
 & Head 4 & 76.91 & 6.34 & 16.08 & 77.43 & 4.70 & 10.49 & 78.42 & 4.05 & 9.02 & 75.71 & 7.04 & 18.88 \\
 & Head 5 & 77.27 & 5.96 & 15.09 & 77.49 & 4.49 & 11.75 & 78.51 & 4.04 & 8.60 & 75.66 & 7.07 & 19.15 \\
    \hline
\multirow{6}{*}{\rotatebox[origin=c]{90}{\begin{tabular}{c}Classifier \\ Ensemble (ours)\end{tabular}}} & Avg. & 77.13 & 5.63 & 13.26 & 77.56 & 4.27 & 9.77 & 78.44 & 3.68 & 8.85 & 75.74 & 6.72 & 17.35 \\
 & Vot. & 76.99 & 5.52 & 13.55 & 77.52 & 4.29 & 10.41 & 78.53 & 3.61 & 8.21 & 75.77 & 6.37 & 15.74 \\
  \cline{2-14} 
 & SL & 76.45 & 2.66 & 7.53 & 77.27 & 2.95 & 11.51 & 78.24 & 2.46 & \textbf{8.05} & 75.09 & \textbf{3.55} & \textbf{7.94} \\
 & DL & 76.27 & \textbf{2.07} & \textbf{5.97} & 77.18 & \textbf{2.69} & 19.08 & 77.66 & 2.23 & 8.88 & 75.36 & 4.44 & 8.68 \\
 & DLL & 76.57 & 3.24 & 9.57 & 77.51 & 2.79 & 11.58 & 78.09 & \textbf{2.19} & 8.86 & 75.43 & 4.84 & 10.36 \\
 & SLpC & 77.22 & 2.80 & 8.99 & 77.39 & 2.95 & \textbf{7.86} & 78.56 & 3.09 & 10.88 & 75.73 & 4.91 & 10.32 \\
    \bottomrule
    \end{tabular}
  \end{footnotesize}
  \rmfamily
\end{table*}

Eight different base models were developed and trained on the CIFAR-100 dataset, with five distinct heads trained for each base model. The results for various ResNet models \cite{he2016deep} are displayed in Table \ref{tab:vergleich-classifier-ensemble-teil1-cifar100}, while more advanced models such as DenseNets \cite{huang2017densely}, ResNeXt \cite{xie2017aggregated} and GoogLeNet \cite{szegedy2015going} are shown in Table \ref{tab:vergleich-classifier-ensemble-teil2-cifar100}. All heads were combined with our classifier ensembles using different methods, including mean averaging, majority voting and different metamodels called Single-Layer (SL), Double-Layer (DL), Double-Layer Large (DLL) and Single-Layer per Class (SLpC). The tables summarize the results for architectures, highlighting accuracy (Acc.), Expected Calibration Error (ECE), and Maximum Calibration Error (MCE).
Individual heads are presented as baseline, where each head paired with the backbone represents a different variation due to the unique random seed applied. The tables present the results for mean averaging and majority voting, followed by the outcomes of the trained metamodels.

\subsubsection{Results for CIFAR-100}\label{sec:cifar}
The results across different ResNet variants on CIFAR-100 (see Table \ref{tab:vergleich-classifier-ensemble-teil1-cifar100}) reveal a consistent trend. For individual heads, the accuracy remains fairly consistent across all ResNet models, with ResNet101 achieving the highest accuracy (78.22\%). However, this also corresponds with higher ECE and MCE values, indicating issues with model calibration. Mean averaging as an ensemble method yields slight improvements in accuracy and moderate reductions in ECE, particularly in ResNet50 and ResNet101, though the calibration improvements are not substantial. Majority voting offers better calibration than mean averaging, resulting in lower ECE and MCE, but accuracy is marginally lower compared to mean averaging.

\begin{table}[t]
  \caption{Comparison of accuracy, ECE and MCE in percent of classic model ensemble with ResNet18 \cite{he2016deep} on CIFAR-100 \cite{krizhevsky2009learning}.}
  \label{tab:vergleich-resnet18-ensemble}
  \renewcommand{\arraystretch}{1.2}
  \centering
  \sffamily
  \begin{footnotesize}
   \begin{tabular}{lrrrr}
    \toprule
\multirow{2}{*}{Model} & \multicolumn{4}{c}{ResNet18} \\
 & Acc. & ECE & MCE  & Params  \\
    \midrule
Model 1 & 74.89 & 5.96 & 27.61  & 11.22 M\\
Model 2 & 75.03 & 6.26 & 17.23 & 11.22 M  \\
Model 3 & 74.20 & 9.72 & 22.02  & 11.22 M  \\
Model 4 & 73.08 & 11.30 & 25.82  & 11.22 M \\
Model 5 & 71.66 & 7.98 & 18.26  & 11.22 M  \\
    \hline
Ensemble & 75.85 & 6.91 & 18.32 & 56.10 M  \\
    \bottomrule
    \end{tabular}
  \end{footnotesize}
  \rmfamily
\end{table}

The metamodels, particularly the SL and DL approaches, show the most significant reductions in ECE and MCE across all ResNet variants, especially for ResNet101. Although these methods slightly decrease accuracy, the calibration improvement is substantial. The DLL and SLpC models also exhibit strong calibration performance, with SLpC performing notably well in terms of ECE for ResNet50.

In line with the findings in Table \ref{tab:vergleich-classifier-ensemble-teil1-cifar100} for ResNet variants, the more advanced models presented in Table \ref{tab:vergleich-classifier-ensemble-teil2-cifar100} display a comparable pattern. While individual heads achieve competitive accuracy, they consistently exhibit higher calibration errors, with GoogLeNet showing particularly elevated ECE and MCE values.

Mean averaging and majority voting marginally reduce calibration errors across all models, particularly in DenseNet and ResNeXt. However, these reductions are not as significant as those seen with the use of metamodels. The trained metamodels, particularly the DL and SL approaches, yield substantial improvements in calibration metrics. The DL metamodel delivers the lowest ECE and MCE values for ResNeXt and DenseNet, with notable performance in reducing calibration errors while maintaining accuracy. The SLpC model also demonstrates good calibration, especially for DenseNet169, which achieves a balance between low ECE and high accuracy.

Compared to the previous table for ResNet models, these results further highlight the effectiveness of classifier ensembles with metamodels in reducing calibration errors, with DL consistently performing well across architectures. However, the trade-off between accuracy and calibration remains present, as seen with the slight dip in accuracy in some metamodel approaches. Overall, classifier ensembles incorporating metamodels continue to significantly enhance model calibration across various architectures, building on the trends observed with the ResNet variants.

As reference, a traditional horizontal model ensemble using ResNet18 was also evaluated. The results are shown in Table \ref{tab:vergleich-resnet18-ensemble}. This approach aggregates models from different checkpoints during training and combines their outputs using mean averaging.
When comparing the ResNet18 results from Table \ref{tab:vergleich-resnet18-ensemble} with those in Table \ref{tab:vergleich-classifier-ensemble-teil1-cifar100}, some distinct trends can be observed. 

The accuracy of the classical model ensemble in Table \ref{tab:vergleich-resnet18-ensemble} reaches 75.85\%, which is slightly higher than the individual heads, where the highest accuracy is 75.25\%. However, the calibration errors, particularly the ECE and MCE, remain relatively high in the classical ensemble, with 6.91\% and 18.32\%, respectively. In contrast, our classifier ensembles using metamodels in Table \ref{tab:vergleich-classifier-ensemble-teil1-cifar100} consistently achieve much lower calibration errors, with the SL and DL approaches reducing the ECE to 2.59\% and 2.93\%, respectively, while also minimizing the MCE.

Another notable difference is the parameter count. The classical ensemble significantly increases the number of parameters to 56.1 M, whereas the classifier ensembles with metamodels only introduce minor increases in parameter count (approximately 3\%). Thus, while the classical ensemble offers slightly improved accuracy, it does so at the cost of significantly higher calibration errors and a substantial increase in model size compared to the classifier ensemble methods.

\subsubsection{Results for Tiny ImageNet}\label{sec:tiny}

\begin{table}[t]
  \caption{Comparison of accuracy, ECE and MCE in percent of uncalibrated heads and their combination to the classifier ensemble with ResNet18 \cite{he2016deep} on Tiny ImageNet \cite{le2015tiny}.}
  \label{tab4:vergleich-classifier-ensemble-teil1-tiny200}
  \renewcommand{\arraystretch}{1.2} 
  \centering
  \sffamily
  \begin{footnotesize}
    \begin{tabular}{llrrr}
    \toprule
\multirow{2}{*}{} & \multirow{2}{*}{Model} & \multicolumn{3}{c}{ResNet18} \\
 & & Acc. & ECE & MCE \\
    \midrule
\multirow{5}{*}{\rotatebox[origin=c]{90}{Baseline}} & Head 1 & 63.41 & 6.02 & 16.56 \\
 & Head 2 & 63.11 & 5.91 & 15.33 \\
 & Head 3 & 63.23 & 6.43 & 18.06 \\
 & Head 4 & 63.39 & 6.05 & 15.89 \\
 & Head 5 & 63.34 & 5.84 & 17.09 \\
    \hline
\multirow{6}{*}{\rotatebox[origin=c]{90}{\begin{tabular}{c}Classifier \\ Ensemble (ours)\end{tabular}}} & Avg. & 63.32 & 5.86 & 16.43 \\
 & Vot. & 63.32 & 5.67 & 14.65 \\
  \cline{2-5} 
 & SL & 62.69 & 5.00 & 11.35 \\
 & DL & 62.05 & 5.09 & 13.19 \\
 & DLL & 62.58 & \textbf{3.13} & \textbf{6.62} \\
 & SLpC & 63.26 & 4.92 & 8.93 \\
    \bottomrule
    \end{tabular}
  \end{footnotesize}
  \rmfamily
\end{table}

Table \ref{tab4:vergleich-classifier-ensemble-teil1-tiny200} presents the results of ResNet18 trained on the Tiny ImageNet dataset, comparing individual heads and various classifier ensemble methods in terms of accuracy (Acc.), Expected Calibration Error (ECE) and Maximum Calibration Error (MCE).

The individual heads achieve accuracy scores around 63.3\%, with ECE values between 5.84\% and 6.43\%, and MCE values ranging from 15.33\% to 18.06\%. The classifier ensemble methods reduce calibration errors, with the DLL approach notably lowering the ECE to 3.13\% and MCE to 6.62\%, significantly outperforming the other methods in terms of calibration. However, the accuracy of the ensemble methods slightly decreases compared to the individual heads.

\section{\uppercase{Conclusions}}
\label{sec:conclusion}

In this study, we explored various ensemble techniques using multiple deep learning architectures on the CIFAR-100 and Tiny ImageNet datasets. Our focus was on evaluating the accuracy and calibration performance of our novel classifier ensembles, particularly in reducing Expected Calibration Error (ECE) and Maximum Calibration Error (MCE). 

The results show that while individual heads achieve reasonable accuracy, they often exhibit high calibration errors, particularly on larger models. Simple ensemble techniques such as mean averaging and majority voting provide modest improvements in calibration but fail to significantly lower the ECE and MCE. In contrast, metamodel-based ensemble methods consistently outperform these basic techniques in terms of calibration, with our Double-Layer and Double-Layer Large methods beeing particularly effective in reducing both ECE and MCE, albeit with slight reductions in accuracy.

Compared to traditional model ensembles, classifier ensembles with metamodels demonstrated similar improvements in calibration with far fewer parameters, offering a more efficient approach to improving model reliability. These findings suggest that integrating metamodels into classifier ensembles can provide a robust solution for enhancing the calibration of deep learning models, making them more reliable in real-world applications.

Future work could explore the scalability of these methods to even larger datasets and architectures, as well as their potential in more complex tasks like object detection requiring highly calibrated predictions.

\section*{\uppercase{Acknowledgements}}
This research was partly funded by Albert and Anneliese Konanz Foundation, the German Research Foundation under grant INST874/9-1 and the Federal Ministry of Education and Research Germany in the project M\textsuperscript{2}Aind-DeepLearning (13FH8I08IA).

\bibliographystyle{apalike}
{\small
\bibliography{main}}

\end{document}